\documentclass[11pt]{article}

\usepackage{epsf}

\def\x{\mathbf{x}}

\def\0{\mathbf{0}}

\def\lam{\lambda}

\title{Polynomial time algorithms for bi-criteria, multi-objective and ratio problems in clustering and imaging\\
Part I: Normalized cut and ratio regions}

\author{Dorit S. Hochbaum
\\
         email: {\tt hochbaum@ieor.berkeley.edu}\\
         }

\newtheorem{thm}{Theorem}[section]

\newcommand{\qed}{\hfill\rule{2mm}{2mm}}

\date{Feb 10, 2008}

\begin{document}
\maketitle

\begin{abstract}
 Partitioning and grouping of similar objects plays a
fundamental role in image segmentation and in clustering problems.  In
such problems a typical goal is to group together similar objects, or
pixels in the case of image processing.  At the same time another goal
is to have each group distinctly dissimilar from the rest and possibly
to have the group size fairly large. These goals are often combined as
a ratio optimization problem. One example of such problem is the
normalized cut problem, another is the ratio regions problem. We devise
here the first polynomial time algorithms solving these problems
optimally. The algorithms are efficient and combinatorial. This
contrasts with the heuristic approaches used in the image segmentation
literature that formulate those problems as nonlinear optimization
problems, which are then relaxed and solved with spectral techniques in
real numbers.  These approaches not only fail to deliver an optimal
solution, but they are also computationally expensive.  The algorithms
presented here use as a subroutine a minimum $s,t$-cut procedure on a
related graph which is of polynomial size. The output consists of the
optimal solution to the respective ratio problem, as well as a sequence
of nested solution with respect to any relative weighting of the
objectives of the numerator and denominator.

An extension of the results here to bi-criteria and multi-criteria
objective functions is presented in part II.
\end{abstract}

\section{Introduction}

The leading challenge in the field of imaging is vision grouping, or
segmentation.  Grouping is to recognize and delineate, automatically,
the salient objects in an image. Image segmentation is equivalent to
partitioning the set of pixels forming the image, or to clustering its
pixels.  High quality clustering is often defined by multiple
attributes.  As an optimization problem this requires attaining more
than one objective.  The motivation for studying the normalized cut
problem is an example of setting an optimization criterion in order to
attain two goals. One goal is to have the selected group's pixels to be
as dissimilar to the remainder of the image as possible, and the second
is to maximize the similarity of the pixels within the group.  These
two objectives are presented as a minimization of the ratio of the
first to the second.

A great deal of the literature is concerned only with the
bipartitioning of the image.  That is, the separation of one object
segment from the rest of the image - the background.   Even this modest
goal presents a number of computational difficulties.   While
presenting an image as a graph and the similarity between pairs of
objects as a weight of an edge, a simple minimum $2$-cut problem will
achieve a partition that minimizes similarity between the two parts. An
adverse phenomenon associated with the minimum $2$-cut optimal
solution, that was noted however by Shi and Malik, \cite{SM}, and
others, is that often the selected part tends to be very small.  To
compensate and correct for the phenomenon of small segments Shi and
Malik introduced the notion of {\em normalized cut}.

Graph theoretical framework is suitable for representing image
segmentation and grouping problems.   The image segmentation problem is
presented on an undirected graph $G=(V,E)$, where $V$ is the set of
pixels and $E$ are the pairs of which similarity information is
available.  Typically one considers a planar image with pixels arranged
along a grid.  The $4$-neighbors set up is commonly used with each
pixel having $4$ neighbors two along the vertical axis and two along
the horizontal axis.  This set up forms a planar grid graph.  The
$8$-neighbors arrangement is also used, but the planar structure is no
longer preserved, and complexity of the various heuristic algorithms is
increasing, sometimes significantly.  Images can of course be also
$3$-dimensional, and in general clustering problems there is no grid
structure.  The algorithms presented here do not assume any specific
property of the graph $G$ - they work for general graphs.

The edges in the graph representing the image carry {\em similarity}
weights.  There is a great deal of literature on how to generate
similarity weights, and we do not discuss this issue here. We only use
the fact that similarity is inversely increasing with the difference in
attributes between the pixels.  In terms of the graph, each edge
$[i,j]$ is assigned a similarity weight $w_{ij}$ that is increasing as
the two pixels $i$ and $j$ are perceived to be more similar.  Low
values of $w_{ij}$ are interpreted as dissimilarity. However, in some
contexts one might want to generate {\em dissimilarity} weights
independently. In that case each edge has two weights, $w_{ij}$ for
similarity, and $\hat{w}_{ij}$ for dissimilarity.

Two applications of efficient algorithms for ratio problems are
presented: One for the problem of {\em ``normalized cut"}, which is to
minimize the ratio of the similarity between the set of objects and its
complement and the similarity within the set of objects.  The second
problem is that of {\em ``ratio-regions"} which is to minimize the
ratio of of the similarity between the set of objects and its
complement and the number (or weight) of the objects within the set.
The algorithms not only provide an optimal solution to the ratio
problem, but also deliver a sequence of solutions for all possible
relative weighting of the two objectives.  These solutions are often
more informative than the optimal solution to the ratio problem alone.

Although the multi-segmentation problem is a partition to multiple
sets, the ratio problems discussed here do not directly addressed as
such, due to computational issues.  Instead these bi-partitions have
been used recursively to generate any desired number of segments.  It
is shown in part II that all the problems presented here have optimal
solutions that are bipartition, and how to characterize ratio problems
in general with this property.  For multiple segments, the Markov
Random Fields is one model that has been popular in that context. It
has been studied by numerous authors, e.g.\ \cite{BZ}, and has been
established for the first time to be polynomial time solvable for
convex objectives in \cite{Hoc01}.

\section{Notation}
Let the weights of the edges in the graph be $w_{ij}$ for $[i,j]\in E$.
If the edges have two sets of weights, these will be denoted by
$w^1_{ij}$ and $w^2_{ij}$.

A bipartition of the graph is called a {\em cut}, $(S,\bar{S})=\{ [i,j]
|i\in S, j\in \bar{S} \}$,  where $\bar{S}=V\setminus S $. We define
the {\em capacity of a cut} $(S,\bar{S})$, and the capacity for any
pair of sets $(A,B)$ to be $C(A,B)= \sum _{i\in A, j\in B} w_{ij}$.  We
define the {\em capacity of a set} $A\subset V$ to be $C(A)= C(A,A)=
\sum _{i,j\in A} w_{ij}$.  For inputs with two sets of edge weights we
let $C_1(A,B)= \sum _{i\in A, j\in B} w^1_{ij}$ and $C_2(A,B)= \sum
_{i\in A, j\in B} w^2_{ij}$.

Given a partition of a graph into $k$ disjoint components, $\{
V_1,\ldots ,V_k\}$ the {\em k-cut} value is $C(V_1,\ldots
,V_k)={\frac{1}{2}}\sum _{i=1}^k C(V_i,\bar{V_i})$.  The problem of
partitioning a graph to $k$ nonempty components that minimize the
$k$-cut value is polynomial time solvable for fixed $k$, \cite{GH}.

For graphs with weighted nodes, we let the weight of node $j$ be $v_j$.
The weight of a set of nodes $A$ is denoted by $V(A)=\sum _{j\in
A}v_j$.

\section{Several ratio problems}
We list here four types of ratio problems.  This include, in addition
to the normalized cut problem and the ratio regions problem, also the
densest subgraph problem and the ``ratio cut" problem.  We solve here
only the first two.  The third problem has been known to be polynomial
time solvable, and the last problem is NP-hard.

\subsection{The normalized cut problem}

Shi and Malik noted in their work on segmentation that cut procedures
tend to create segments that may be very small in size.  To address
this issue they proposed several versions of objective functions that
provide larger segments in an optimal solution.  Among the proposed
objective they formulated the normalized cut as the optimization
problem
$$\min _{S\subset V} C(S,\bar{S})\cdot (\frac{1}{|S|} +
\frac{1}{|\bar{S}|} ).$$

This problem is equivalent to finding the expander ratio of the graph
discussed in the next subsection.

This objective function drives the segment $S$ and its complement to be
approximately of equal size.  Indeed, like the balanced cut problem the
problem was shown to be NP-hard, \cite{SM}, by reduction from set
partitioning. A variant of the problem also defined by Shi and Malik is
$$\min _{S\subset V} C(S,\bar{S})\cdot (\frac{1}{C(S,V)} +
\frac{1}{C(\bar{S},V)} ).$$

Another variant yet of the problem is the quantity $h_G = \min
\frac{C(S,V\setminus S)}{\min \{ C(S,S),C(V\setminus S)\}}$, also known
as the {\em Cheeger constant}, \cite{Chee,Chung}. More frequently, for
minor variants of the problem, the denominator is $\min \{
|S|,|V\setminus S|\}$, or $|S|$ is replaced by a quantity representing
the volume of $S$. This Cheeger constant is approximated by the second
largest eigenvalue of a certain related adjacency matrix of the graph.
This eigenvalue $\lambda _1$ is related to the Cheeger constant by the
inequalities: $2h_G \geq \lambda _1 \geq h_G^2/ 2 $. Computing the
value of the Cheeger constant is NP-hard - it is the same as finding
the expander ratio of a graph and again it drives to a roughly equal or
balanced partition of the graph. The dominant techniques in vision
grouping are spectral in nature. That is, they compute the eigenvalues
and the eigenvectors and then some type of rounding process, see e.g.\
\cite{TM,SGS}.

Instead of the sum problem, there are other related optimization
problems used for image segmentation.  Sharon et al. \cite{SGS} define
the normalized cut as
$$\min _{S\subset V} \frac{C(S,\bar{S})}{C(S,S)}.$$

Sharon et al. \cite{SGS} state that:
\begin{quote}
A salient segment in the image is one for which the similarity across
its border is small, whereas the similarity within the segment is large
(for a mathematical description, see Methods). We can thus seek a
segment that minimizes the ratio of these two expressions. Despite its
conceptual usefulness, minimizing this ænormalized cutÆ measure is
computationally prohibitive, with cost that increases exponentially
with image size.
\end{quote}

One of our contributions here is to show that the problem of minimizing
this ratio is in fact solvable in polynomial time, and with a
combinatorial algorithm.

The typical solution approach used when addressing optimization
problems for image segmentation is to approximate the problem objective
by a nonlinear (quadratic) expression for which the eigenvectors of an
associated matrix form an optimal solution.  Let binary variables $x_i$
for $i\in V$ be defined so that $x_i=1$ if node $i$ in the selected
side of the cut -- the segment. The following nonlinear formulation is
the relaxation that has been used by Sharon et al.\ and others,
\cite{SGS,TM,SM}:

$$\min \frac{\sum w_{ij}(x_i-x_j)^2}{\sum w_{ij} x_i\cdot x_j}=\frac {\x ^T L \x}{\x^T W \x },$$

where $L$ is the Laplacian matrix of the graph and $W$ is a matrix
appropriately defined.   The use of spectral techniques involves real
number computations with the associated numerical issues.  Even an
exact solution to the nonlinear problem is a vector of real numbers
whereas the original problem is discrete and binary.

However, this normalized cut problem (without the ``balanced"
requirement) is polynomial time solvable.  We show an algorithm solving
the problem in the same complexity as a single minimum $s,t$-cut on a
related graph on $O(n+m)$ nodes and $O(n+m)$ edges.

\subsection{Ratio regions and expanders} Consider the objective
function  $\min _{|S|\leq {n\over 2}} \frac{C(S,\bar{S})}{|S|}$.  This
value of the optimal solution, for a graph $G$, is known as the
expansion ratio of $G$.  This problem is NP-hard as the limit on the
size of $|S|$ makes it difficult, and drives the solution towards a
{\em balanced} cut -- a known NP-hard problem. The objective function
can also be written as $\min _{S\subset V} \frac{C(S,\bar{S})}{\min
\{|S|,|\bar{S}| \}}$.

A variant of this problem $\min _{S\subset V} \frac{C(S,\bar{S})}{|S|}$
has been considered under the name {\em ratio regions} by Cox et al.\
\cite{Cox}.  The ratio region problem is motivated by seeking a
segment, or region, where the boundary is of low cost and the segment
itself has high weight.   The problem studied by Cox et al.\ is
restricted to planar graphs and thus to planar grid images with $4$
neighbors only. Though the boundary of the region is defined as a path,
the path corresponds to a cut in the dual graph. For graph nodes of
weight $d_i$ the problem is $\min _{S\subset V}
\frac{C(S,\bar{S})}{\sum _{i\in S}d_i}$.

This problem is shown here to be polynomially solvable by a parametric
cut procedure, in the complexity of a single minimum cut. The problem
is in fact equivalent to a binary and linear version of the Markov
Random Fields problem, called the maximum $s$-excess problem in
\cite{Hoc97}. It is interesting to note that the pseudoflow algorithm
in \cite{Hoc97} is set to solve the maximum $s$-excess problem
directly.  Our algorithm for the ratio regions problem applies for node
weights that can be either positive or negative.  This generalizes the
application context of Cox et al.\ the node weighs were all positive.

\subsection{Densest subgraph}
Sarkar and Boyer \cite{SB} defined the problem $\min _{S\subset V}
\frac{C(S,S)}{|S|}$. This objective is of interest for weights that
reflect dissimilarity.  In that case the goal is to minimize the
dissimilarity within the selected segment while the size of the segment
tends to be large.   For similarity weights the objective would be to
maximize this quantity. Both this problem, and its maximization version
are solved in polynomial time.  Also the nodes can carry arbitrary
weights and the algorithm is still applicable with no change in the
running time.

This problem has been known for a long time as the maximum density
subgraph is the subgraph induced by the subset of nodes $D$ maximizing,

$$\max _{S\subset V} \frac{C(S,{S})}{|S|}.$$

This problem was shown to be solvable in polynomial time by Goldberg
\cite{Gol}. Gallo, Grigoriadis and Tarjan \cite{GGT} showed how the
problem would be solved as a parametric minimum $s,t$-cut in the
complexity of a single $s,t$-cut.

A node weighted version of the problem is $\max _{S\subset V}
\frac{C(S,{S})}{V(S)}.$  This problem is solved by a minor extension of
the densest subgraph approach in the same run time.

\subsection{``Ratio cuts"}
This problem was introduced by Wang and Siskind \cite{Sis}.  In the
ratio cut problem the edges have two weights associated with each.
Wang and Siskind studied the case where $w^1_{ij}$ are positive and
$w^2_{ij}$ are equal to $1$ for all $[i,j]\in E$.   The goal is to
minimize the ratio,
$$\min _{S\subset V} \frac{C_1(S,\bar{S})}{C_2(S,\bar{S})}.$$

This problem was shown in \cite{Sis} to be at least as hard as the
sparsest cut problem, and therefore NP-hard.  On the other hand, for
planar graphs, Wang and Siskind demonstrated that the problem is
solvable in polynomial time.

\subsection{Overview}

The problems, that the methodology paradigm presented here solve in
polynomial time, are summarized in the Table \ref{tab:1}:

\begin{table} [htb]
\begin{center} \begin{footnotesize}

\begin{tabular}{|l|l|l|} \hline
{\bf Problem name} & {\bf Objective} & {\bf  Reference}
\\ \hline \hline
Normalized cut & $\min _{S\subset V} \frac{C(S,\bar{S})}{C(S,S)}$ &  \cite{SM,SGS}\\
Normalized cut' & $\min \frac{C(S,\bar{S})}{C(S,V)}$ & \cite{SM}\\
``Density" & $\min \frac{C(S,S)}{|S|}$ &  \cite{SB}\\
Ratio regions & $\min \frac{C(S,\bar{S})}{|S|}$ & \cite{Cox}\\
Weighted ratio regions & $\min \frac{C(S,\bar{S})}{\sum _{i\in S} d_i}$ & \cite{Cox}\\
\hline \hline
\end{tabular}
\caption{Ratio optimization problems in image segmentation}
\label{tab:1}
\end{footnotesize}
\end{center}
\end{table}

We present here in detail only the algorithm for the normalized cut
problem, which is the hardest one on the list. The construction for the
ratio regions is described briefly.  For the ``density" problem with
either a minimization or maximization objective function we studied the
problem for the entire sequence of solutions in the context of
dynamically evolving facility set, see \cite{Hoc05}.  Thus for the
"density" case we only point out an efficient algorithm.

\section{The solution approach}
\subsection{Monotone integer programming formulation}
The key is to formulate the problem as an integer linear programming
problem, a 0-1 integer programming here, with {\em monotone}
inequalities constraints.  It was shown in \cite{Hoc02} that any
integer programming formulation on monotone constraints has a
corresponding graph where the minimum cut solution corresponds to the
optimal solution to the integer programming problem.  Thus the
formulation is solvable in polynomial time.

To convert the ratio objective to a linear objective we utilize the
reduction of the ratio problem to a linearized optimization problem.

\subsection{Linearizing ratio problems}
 A general approach for
maximizing a fractional (or as it is sometimes called, geometric)
objective function over a feasible region ${\cal{F}}$, $\min _{x\in
{\cal{F}}} \frac{f(x)}{g(x)}$, is to reduce it to a sequence of calls
to an oracle that provides the yes/no answer to the $\lambda$-{\em
question}:\\ Is there a feasible subset $V'\subset V$ such that $\sum
_{x\in {\cal{F}}} f(x) - \lambda \sum _{x\in {\cal{F}}} g(x) <0 $?\\

 If the answer to the $\lambda$-question is {\em yes} then the optimal
solution has value smaller than $\lambda$.   Otherwise, the optimal
value is greater than or equal to $\lambda$.  A standard approach is
then to utilize a binary search procedure that calls for the
$\lambda$-question $O(\log (UF))$ times in order to solve the problem,
where $U=\sum _{[i,j]\in E}w_{ij}$, and $F=\sum_{[i,j]\in E}w'_{ij}$
for the weights at the denominator $w'_{ij}$.

Therefore, if the linearized version of the problem, that is the
$\lambda$-question, is solved in polynomial time, then so is the ratio
problem.   Note that the number of calls to the linear optimization is
not strongly polynomial but rather, if binary search is employed,
depends on the logarithm of the magnitude of the numbers in the input.
In some cases however there is an efficient procedure that uses the
solution for one parameter value to compute the value for another
parameter value more efficiently.  Such is the case for the densest
subgraph problem which has an efficient parametric procedure
(\cite{GGT}).

It is important to note that {\em not all} ratio problems are solvable
in polynomial time.  One prominent example of such ratio problem is the
ratio cut introduced by Wang and Siskind, \cite{Sis}.  That criterion
applies in a graph with two sets of weights for each edge. Let $w^1
_{ij}$, $w^2 _{ij}$ be the two weights assigned to edge $[i,j]$.  Then
a cut with respect to $w^1 _{ij}$ separating $S$ from $\bar{S}$ is $C_1
(S,\bar{S})$ and with respect to $w^2 _{ij}$ is is denoted by $C_2
(S,\bar{S})$.  The ratio criterion defined by Wang and Siskind is to
minimize $\frac{C_1 (S,\bar{S})}{C_2(S,\bar{S})}$. For the weight $w^2
_{ij}$ they use the value $1$, so this objective is to find a cut
minimizing the cut value divided by the number of edges in the cut.  As
in other cases, the rationale is to try and increase the number of
edges in the cut, and hence the size of the cluster/segment. This
particular criterion was shown in \cite{Sis} to be NP-hard, and
polynomial time solvable on planar graphs. For the ratio cut problem
the linearized problem is NP-hard, and the ratio cut problem is NP-hard
as well.  The linearized problem is NP-hard by reduction from maximum
cut, and the ratio problem by reduction from the sparsest cut problem,
\cite{Sis}. However the ratio cut problem has a polynomial time
algorithm for planar graphs. For planar graphs the $\lambda$-question
is solved by finding a maximum weight non-bipartite matching in a
related graph. The procedure of \cite{Sis} indeed makes repeated calls
to solving non-bipartite matching problems, where for each value of
$\lambda$ another graph has to be constructed.

It is also important to note that linearizing is not always the right
approach to use for a ratio problem.  For example, the problem of
finding a partition of a graph to $k$ components minimizing the $k$-cut
between components for $k\geq 2$ divided by the number of components
$k$, always has an optimal solution with $k=2$ which is attained by a
minimum $2$-cut algorithm. On the other hand, the linearized problem is
much harder to solve (though it can be solved in polynomial time.)
Additional details are provided in part II of this paper.

\section{The normalized cut formulation}

We first provide a formulation for the problem, $\min _{S\subset V}
\frac{C_1(S,\bar{S})}{C_2(S,S)}.$   We use different similarity weights
for the numerator $w_{ij}'$ and denominator $w_{ij}$.

We begin with an integer programming formulation of the problem. Let
\[ x_i= \left\{ \begin{array}{ll}
      1 & \mbox{ if     }\  i\in S \\
      0 & \mbox{ if     }\  i\in \bar{S} .
                \end{array}
\right. \]
We define additional binary variables: $z_{ij}=1$ if exactly
one of $i$ or $j$ is in $S$; $y_{ij}=1$ if both $i$ or $j$ are in $S$.

\[ z_{ij}= \left\{ \begin{array}{ll}
      1 & \mbox{ if     }\  i\in S, j\in \bar{S}, \mbox { or  } i\in \bar{S}, j\in S  \\
      0 & \mbox{ if     }\  i,j\in S \mbox { or  } i,j\in  \bar{S} .
                \end{array}
\right. \]

\[ y_{ij}= \left\{ \begin{array}{ll}
      1 & \mbox{ if     }\  i,j\in S \mbox{ or } i,j\in \bar{S}\\
      0 & \mbox{ otherwise.     }
                \end{array}
\right. \]

With these variables the following is a valid formulation (NC) of the
normalized cut problem:

\[
\hspace{.4in}\begin{array}{ll} \mbox{(NC)~~} \min \ & \frac{\sum
w_{ij}z_{ij}}{\sum
w'_{ij}y_{ij}}\\
\mbox{subject to } \ & x_{ i} - x_{j}  \leq z_{ij}  \quad
\mbox{for all $[i,j]\in E$ } \\
\ & x_{j} - x_{i}  \leq   z_{ji}\quad   \mbox{for all $[i,j]\in E$ }\\
\ &  y_{ij} \leq x_i  \quad   \mbox{for all $[i,j]\in E$ }\\
\ &  y_{ij} \leq x_{j}\\
\ & 1\leq \sum _{[i,j]\in E} y_{ij} \leq |E|-1 \\
   & x_j {\rm \ binary \ } j\in V\\
   &   z_{ij},y_{ij} {\rm \ binary \ } j\in V.
\end{array}
\]

To verify the validity of the formulation notice that the objective
function drives the values of $z_{ij}$ to be as small as possible, and
the values of $y_{ij}$ to be as large as possible.  With the
constraints, $z_{ij}$ cannot be $0$ unless both endpoints $i$ and $j$
are in the same set.  On the other hand $y_{ij}$ cannot be equal to $1$
unless both endpoints $i$ and $j$ are in $S$.  The sum constraint
ensures that at least one edge is in the segment $S$ and at least one
edge is in the complement - the background. Otherwise the ratio is not
be defined in the first case and the optimal solution is to choose
$S=V$ in the second.   We remove the sum constraint from the
formulation and replace it by setting for some edge in the object to
serve as ``seed" and some edge in the background to serve as ``seed".
Adding seeds is an approach often used in segmentation, see e.g.\
\cite{BJ} where the user keeps adding ``seeds" until the bipartition is
satisfactory.  Here the choice is made once, and can be replaced by
enumerating the possible pairs of edges that serve as object and
background edges.  Since for both the object and it complement the cut
value is the same, the solution will always be in terms of the larger
segment in the bipartition that is likely to contain higher total
similarity weights.  The edge that we set to be in the source is
therefore usually the one in the background and the one in the sink
would be in the object. We thus replace the sum constraint by setting
$y_{i^*j^*}=1$ and $y_{i'j'}=0$ for some pair of edges in the
background and the object respectively.

Once the sum constraint has been removed, the problem formulation is
easily recognized as a monotone integer programming with up to three
variables per inequality according to the definition provided in
Hochbaum's \cite{Hoc02}. Any such problem was shown there to be
solvable as a minimum cut problem on a certain associated graph.
Because the objective function is a ratio, we first ``linearize" the
problem.

\subsection{Linearizing the objective function}
The $\lambda$-{\em question} for the normalized cut problem is:\\ Is
there a feasible subset $V'\subset V$ such that $\sum
_{[i,j]\in E} w_{ij}z_{ij} - \lambda \sum _{[i,j]\in E} w'_{ij}y_{ij} <0 $?\\

One possible approach is to utilize a binary search procedure that
calls for the $\lambda$-question $O(\log (UF))$ times in order to solve
the problem, where $U=\sum _{[i,j]\in E}w_{ij}$, and $F=\sum_{[i,j]\in
E}w'_{ij}$ for the weights at the denominator $w'_{ij}$.

With the construction of the graph we observe that one can use instead
a parametric approach which is significantly more efficient.  We note
that the $\lambda$-question is the following {\em monotone}
optimization problem,

\[
\hspace{.4in}\begin{array}{ll} \mbox{($\lambda$-NC)~~}  \min \ &\sum
_{[i,j]\in E} w_{ij}z_{ij} -  \sum _{[i,j]\in E} \lambda w'_{ij}y_{ij}\\
\mbox{subject to } \ & x_{ i} - x_{j}  \leq z_{ij}  \quad
\mbox{for all $[i,j]\in E$ } \\
\ & x_{j} - x_{i}  \leq   z_{ji}\quad   \mbox{for all $[i,j]\in E$ }\\
\ &  y_{ij} \leq x_i  \quad   \mbox{for all $[i,j]\in E$ }\\
\ &  y_{ij} \leq x_{j}\\
\ & y_{i^*j^*}=1 \mbox{  and } y_{i'j'}=0\\
   & x_j {\rm \ binary \ } j\in V\\
   &   z_{ij},y_{ij} {\rm \ binary \ } j\in V.
\end{array}
\]

If the optimal value of this problem is negative then the answer is
yes, otherwise the answer is no.  This problem is an integer
optimization problem on a totally unimodular constraint matrix. That
means that we can solve the linear programming relaxation of this
problem and get a basic optimal solution that is integer.  Instead we
will use a much more efficient algorithm described in \cite{Hoc02}
which relies on the monotone property of the constraints.

\subsection{Solving the $\lambda$ question with a minimum cut procedure}

\begin{figure}[thb]
\vspace{-1.5in} \epsfxsize = 1.2\linewidth
\centerline{\hspace{-0.75in}\epsfbox{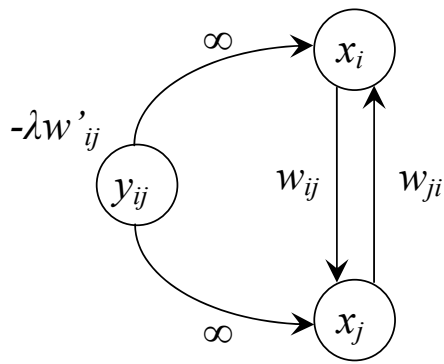}} \vspace{-6.2in}
\caption{\label{fig:1}The basic gadget in the graph representation.}
\end{figure}

We construct a directed graph $G'=(V',A')$ with a set of nodes $V'$
that has a node for each variable $x_i$ and a node for each variable
$y_{ij}$. The nodes $y_{ij}$ carry a negative weight of $-\lambda
w_{ij}$.  The arc from $x_i$ to $x_j$ has capacity $w'_{ij}$ and so
does the arc from $x_j$ to $x_i$ as in our problem $w_{ij}=w_{ji}$. The
two arcs from each edge-node $y_{ij}$ to the endpoint nodes $x_i$ and
$x_j$ have infinite capacity. Figure \ref{fig:1} shows the basic gadget
in the graph $G'$ for each edge $[i,j]\in E$.

We claim that any finite cut in this graph, that has $y_{i^*j^*}$ on
one side of the bipartition and $y_{i'j'}$ on the other, corresponds to
a feasible solution to the problem $\lambda$-NC.  Let the cut $(S,T)$,
where $T=V'\setminus S$, be of finite capacity $C(S,T)$.  We set the
value of the variable $x_i$ or $y_{ij}$ to be equal to $1$ if the
corresponding node is in $S$, and $0$ otherwise.  Because the cut is
finite, then $y_{ij}=1$ implies that $x_{i}=1$ and $x_j=1$.

Next we claim that for any finite cut the sum of the weights of the
$y_{ij}$ nodes in the source set and the capacity of the cut is equal
to the objective value of problem $\lambda$-NC.  Notice that if
$x_{i}=1$ and $x_j=0$ then the arc from the node $x_{i}$ to node $x_j$
is in the cut and therefore the value of $z_{ij}$ is equal to $1$.

\begin{figure}[thb]
\vspace{-0.8in} \epsfxsize = 0.8\linewidth
\centerline{\hspace{-0.6in}\epsfbox{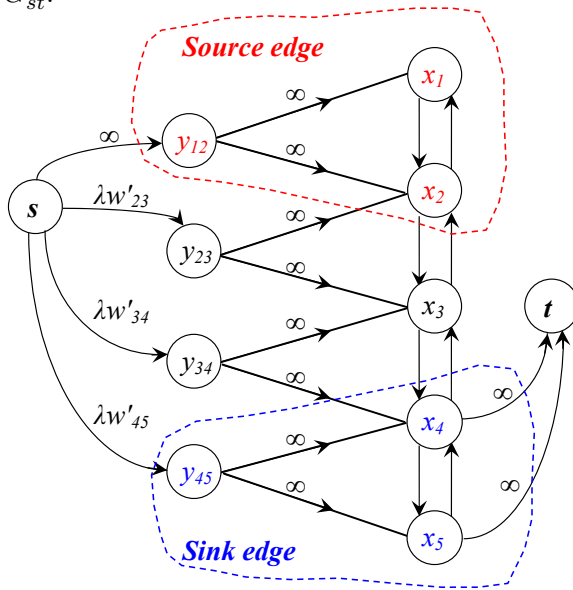}} \vspace{-2.8in}
\caption{The graph $G'_{st}$ with edge $[1,2]$ as source seed and edge
$[4,5]$ as sink seed. \label{fig:3}}
\end{figure}

We next create a source node $s$ and connect all $y_{ij}$ nodes to the
source with arcs of capacity $\lambda w'_{ij}$.  The node $ y_{i^*j^*}$
is then shrunk with a source node $s$ and therefore also its endpoints
nodes. The node $ y_{i'j'}$ and its endpoints nodes are shrunk with the
sink $t$.  We denote this graph illustrated in Figure \ref{fig:3},
$G'_{st}$.

\begin{thm}
A minimum $s,t$-cut in the graph $G'_{st}$, $(S,T)$, corresponds to an
optimal solution to $\lambda$-NC by setting all the variables whose
nodes fall in $S$ to $1$ and zero otherwise.
\end{thm}
{\bf Proof:} Note that whenever a node $y_{ij}$ is in the sink set $T$
the arc connecting it to the source is included in the cut. Let the set
of $x$ variable nodes be denoted by $V_x$ and the set of $y$ variable
nodes, excluding $ y_{i^*j^*}$, be denoted by $V_y$. Let $(S,T)$ be any
finite cut in $G'_{st}$ with $s\in S$ and $t\in T$ and capacity
$C(S,T)$.
\begin{eqnarray*}
C(S,T) & = &
 \sum_{ y_{ij} \in T \cap V_y}\lambda w'_{ij} +\sum _{i\in V_x\cap S,j\in V_x\cap T}w_{ij} \\
{} & = & \sum_{ v \in V_y}\lambda w'_{v} - \sum_{ y_{ij} \in S \cap
V_y}\lambda w'_{ij} +\sum _{x_i\in V_x\cap S,x_j\in V_x\cap T}w_{ij} \\
{} & = & \lambda W' + [\sum _{i\in V_x\cap S,j\in V_x\cap T}w_{ij}-
\sum_{ y_{ij} \in S \cap V_y}\lambda w'_{ij} ]. \\
\end{eqnarray*}

This proves that for a fixed constant $W'=\sum_{ v \in V_y} w'_{v}$ the
capacity of a cut is equal to a constant $W'\lambda$ plus the objective
value corresponding to the feasible solution.  Hence the partition
$(S,T)$ minimizing the capacity of the cut minimizes also the objective
function of $\lambda$-NC. \qed

\subsection{A parametric procedure for solving normalized cut}
The source adjacent arcs in the graph $G'_{st}$ are monotone increasing
with $\lambda$.  As the value of $\lambda$ increases the source set of
the respective minimum cuts are nested.  This is called the nestedness
lemma.   Although the capacity of the cut is increasing with any
increase in $\lambda$ the set of nodes in the source set can thus be
incremented only $n'=|V'|$ times. We call the values of $\lambda$ where
the source set expands by at least one node, the {\em breakpoints} of
the parametric cut.  Let the breakpoints be $\lambda _1 > \lambda _2
>\ldots > \lambda _{\ell}$, with corresponding minimal source sets,
$S_1\subset S_2\subset \ldots \subset S_{\ell}$. As a result of the
nestedness lemma $\ell \leq n'$ for a graph on $n'$ nodes since there
can be no more than $n'$ different nested source sets.  The capacity
value of the minimum cut is increasing as a function of increasing
values of $\lambda$ along a piecewise linear concave curve.

\begin{thm}
All breakpoints of the density graph can be found by solving a
parametric minimum cut problem where the source adjacent capacities of
arcs are linear functions of the parameter, $\lambda$.
\end{thm}

Gallo, Grigoriadis and Tarjan showed in \cite{GGT} how to find all the
breakpoints and the corresponding minimum cuts in the same complexity
as that required to solve a single minimum $s,t$-cut problem with the
push-relabel algorithm of \cite{GT88}.   The pseudoflow algorithm for
maximum flow and minimum cut (see Hochbaum \cite{Hoc97}) also finds the
parametric breakpoints in the complexity of a single minimum $s,t$-cut.
Once all the breakpoints are generated, we search for the largest value
of $\lambda$ among the breakpoints so that the optimal value of
$\lambda$-NC is negative, or equivalently, the minimum $s,t$-cut value
that is strictly less than $\lambda W'$.

To summarize, let $T(n,m)$ be the running time required to solve the
minimum cut problem on a graph with $n$ nodes and $m$ arcs.  In the
graph $G'_{st}$ the number of nodes is $n'=n+m$ where $m$ is the number
of adjacencies  or edges in the image graph. The number of arcs in
$G'_{st}$, $m'=|A'|$, is $O(m)$. For general graph this running time is
$O(m^2\log m)$ with either the pseudoflow algorithm or the push-relabel
algorithm.  The degree of each node is constant for imaging
applications so for that context $m'=O(n)$ and $n'$ is $O(n)$ and the
running time is $O(n^2\log n)$.

\begin{thm}
The normalized cut problem is solvable in the running time of a minimum
$s,t$-cut problem, $T(n',m')$.
\end{thm}

\noindent {\bf Remark:} It may be desirable to solve ($\lambda$-NC)
without specifying a source and a sink.  The problem is then to
partition the graph $G'_{st}$ to two nonempty components so that the
cut separating them is minimum. This problem is the directed minimum
$2$-cut problem. It was shown by Hao and Orlin \cite{HO}, that the
directed minimum $2$-cut problem is solved in the same complexity as a
single minimum $s,t$ cut problem, with the push-relabel algorithm.
(This was shown to hold also for the pseudoflow algorithm.)

In order to solve the normalized cut problem, the algorithm produces a
sequence of nested solutions for all possible values of the parameter
$\lambda$.  Each such solution represents a different weighting of the
cut objective versus the similarity objective.  As the value of the
$\lambda$ grows the similarity objective is more prominent and the
optimal solution $S_{\lambda}$ expands.  Although the normalized cut
ratio problem's optimal solution comprises of a single connected
component (see part II), the sequence of optimal solutions to the range
of parameter values is not necessarily formed of a single connected
component.  Such solutions could be more meaningful in medical images
for instance, where lesions are the features sought, but they often
appear as disjoint components in the image.

\section{A sketch of the technique for Ratio Regions}
The weighted ratio regions problem, once linearized, is an instance of
the $s$-excess problem in \cite{Hoc97}.

As before we formulate the problem first. Let
\[ x_i= \left\{ \begin{array}{ll}
      1 & \mbox{ if     }\  i\in S \\
      0 & \mbox{ if     }\  i\in \bar{S} .
                \end{array}
\right. \] Let $z_{ij}=1$ if exactly one of $i$ or $j$ is in $S$.

\[ z_{ij}= \left\{ \begin{array}{ll}
      1 & \mbox{ if     }\  i\in S, j\in \bar{S}, \mbox { or  } i\in \bar{S}, j\in S  \\
      0 & \mbox{ if     }\  i,j\in S \mbox { or  } i,j\in  \bar{S} .
                \end{array}
\right. \]

Let the similarity weight on each edge be $w_{ij}$ and the weight of
node (pixel) $j$ be $v_j$.  With these parameters the ratio regions
problem formulation is,

\[
\hspace{.4in}\begin{array}{ll} \mbox{(RR)~~} \min \ & \frac{\sum
w_{ij}z_{ij}}{\sum
v_j x_j}\\
\mbox{subject to } \ & x_{ i} - x_{j}  \leq z_{ij}  \quad
\mbox{for all $[i,j]\in E$ } \\
\ & x_{j} - x_{i}  \leq   z_{ji}\quad   \mbox{for all $[i,j]\in E$ }\\
   & x_j {\rm \ binary \ } j\in V\\
   &   z_{ij} {\rm \ binary \ } j\in V.
\end{array}
\]

The corresponding $\lambda$-question is,
\[
\hspace{.4in}\begin{array}{ll} \mbox{($\lambda$-RR)~~}  \min \ &\sum
_{[i,j]\in E} w_{ij}z_{ij} -  \sum _{j\in V} \lambda
v_j x_j\\
\mbox{subject to } \ & x_{ i} - x_{j}  \leq z_{ij}  \quad
\mbox{for all $[i,j]\in E$ } \\
\ & x_{j} - x_{i}  \leq   z_{ji}\quad   \mbox{for all $[i,j]\in E$ }\\
   & x_j {\rm \ binary \ } j\in V\\
   &   z_{ij} {\rm \ binary \ } j\in V.
\end{array}
\]

The graph constructed for that problem is of the same size as the
original graph $G$.  Each node representing a variable $x_j$ has an arc
going to sink node with capacity $\lambda v_j$.  One variable node,
$x_s$ is selected arbitrarily as corresponding to a source ``seed".

\begin{figure}[thb]
\vspace{-0.95in} \epsfxsize = 0.85\linewidth
\centerline{\hspace{0.3in}\epsfbox{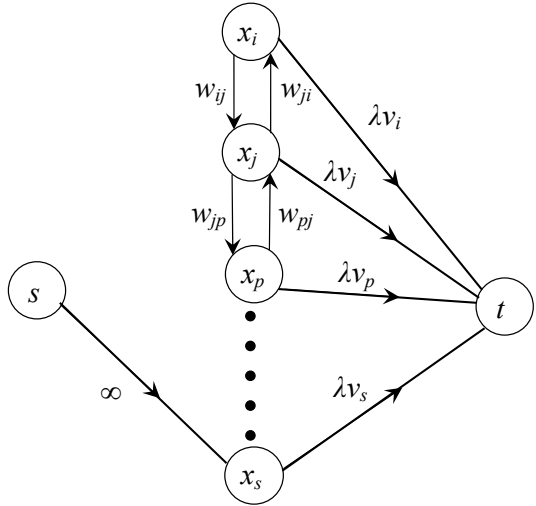}} \vspace{-3.4in}
\caption{The graph $G'_{st}$ for the ratio regions problem with node
$x_s$ serving as source seed. \label{fig:2}}
\end{figure}

 The graph $G'_{st}$  has $O(n)$ nodes and $O(m)$ arcs.  The
algorithm solving the problem is then a simple parametric cut algorithm
in that graph, with run time $T(n,m)$.   Furthermore, the parametric
$s,t$-cut algorithm delivers the sequence of optimal nested solutions
for all values of $\lambda$, as well as the optimal solution to the
ratio problem, in run time $T(n,m)$.

\section{Brief remarks about the other problems}
The problem normalized cut' is: $\min _{S\subset V}
\frac{C(S,\bar{S})}{C(S,V)}$.  This problem is in fact identical to the
normalized cut problem.  To see that notice that
$C(S,V)=C(S,S)+C(S,\bar{S})$.  Substituting this we get:
\begin{eqnarray*}
\frac{C(S,\bar{S})}{C(S,V)} & = &
\frac{C(S,\bar{S})}{C(S,S)+C(S,\bar{S})} = \frac{1}{1+
\frac{C(S,S)}{C(S,\bar{S})}} .
\end{eqnarray*}
This quantity is minimized when $\frac{C(S,S)}{C(S,\bar{S})}$ is
maximized. Thus the same algorithm applies.

Concerning the maximum density problem, it is presented as a minimum
$s,t$-cut problem on an unbalanced bipartite graph with nodes
representing the edges of the graph $G$ on one side of the bipartition
and nodes representing the nodes of $G$ on the other.  (Details are
available in \cite{Hoc05}.)  That bipartite graph has $m+n$ nodes, and
$m'=O(m)$ arcs.  The complexity of a single minimum $s,t$-cut in such
graph is therefore $O(m^2\log m)$.  This however can be improved.

The number of iterations required by the push-relabel algorithm or the
pseudoflow algorithm is bounded by a function of the length of the
longest residual path in the graph  -- $O(m'n')$ -- where $m'$ is the
number of arcs in the bipartite graph and $n'$ is the maximum residual
path length. In the $\lam$-network constructed for the $\lam$-question,
this length $n'$ is at most $2n+2$ as each path alternates between the
two sets in the partition.

This fact is used by Ahuja, Orlin, Stein and Tarjan, \cite{AOST}, who
devised improved push-relabel algorithms for unbalanced bipartite
graphs. Among those, the most efficient for parametric minimum cut is
an adaptation of the parametric push-relabel algorithm of Gallo,
Grigoriadis and Tarjan with run time $O(m'n'\log (\frac{n'^2}{m'}+2))$.
This run time translates to $O(mn\log (\frac{n^2}{m}+2))$ for the
parametric problem solving the minimum density problem on a general
graph, improving on the $O(m^2\log m)$ complexity for a direct
application of the parametric cut algorithm.

\begin{figure}[thb!]
\vspace{-0.4in} \epsfxsize = 0.8\linewidth
\centerline{\epsfbox{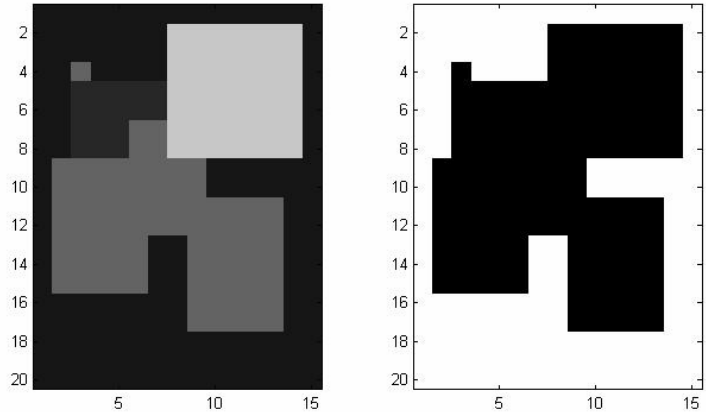}} \vspace{-0.6in} \caption{The
input to the normalized cut procedure on the left.  The output, on the
right, segmented and separated the feature from the
background\label{fig:NC}.}
\end{figure}

\section{Experimental example}
The normalized cut procedure was implemented using the pseudoflow
algorithm, \cite{Hoc97}, which solves the minimum $s,t$-cut problem
(and the maximum flow problem.)  The pseudoflow algorithm is fast in
theory and in practice, \cite{CH}.  The code and its parametric version
are available for download at
http://riot.ieor.berkeley.edu/riot/Applications/Pseudoflow/.

The algorithm described here was applied to a synthetic image provided
in Figure \ref{fig:NC}. The optimal solution differentiated precisely
the feature from the background.

\end{document}